\documentclass[runningheads]{llncs}
\usepackage{graphicx}
\begin{document}
\begin{sloppypar}
%
\title{Impact of Light and Shadow on Robustness of Deep Neural Networks}
%
%
\author{Chengyin Hu\inst{1,2} \and
Weiwen Shi\inst{1} \and
Chao Li\inst{2, 3} \and
Jialiang Sun\inst{2} \and
Donghua Wang\inst{2, 5} \and
Junqi Wu\inst{2, 4} \and
Guijian Tang\inst{2, 6} }

\authorrunning{Chengyin Hu et al.}
%
\institute{University of Electronic Science and Technology of China, Chengdu, China \and
Chinese Academy of Military Science, Beijing, China
\and Xi'an University of Electronic Science and Technology, Xi'an, China
\and Shanghai Jiaotong University, Shanghai, China
\and Zhejiang University, Hangzhou, China
\and National University of Defense Technology, Changsha, China\\
\email{cyhuuestc@gmail.com, weiwen\_shi@foxmail.com, lichaoedu@126.com, sun1903676706@163.com, wangdonghua@zju.edu.cn, wujq28@sjtu.edu.cn, tangbanllniu@163.com}\\
}

\maketitle              
\begin{abstract}
Deep neural networks (DNNs) have made remarkable strides in various computer vision tasks, including image classification, segmentation, and object detection. However, recent research has revealed a vulnerability in advanced DNNs when faced with deliberate manipulations of input data, known as adversarial attacks. Moreover, the accuracy of DNNs is heavily influenced by the distribution of the training dataset. Distortions or perturbations in the color space of input images can introduce out-of-distribution data, resulting in misclassification. In this work, we propose a brightness-variation dataset, which incorporates 24 distinct brightness levels for each image within a subset of ImageNet. This dataset enables us to simulate the effects of light and shadow on the images, so as is to investigate the impact of light and shadow on the performance of DNNs. In our study, we conduct experiments using several state-of-the-art DNN architectures on the aforementioned dataset. Through our analysis, we discover a noteworthy positive correlation between the brightness levels and the loss of accuracy in DNNs. Furthermore, we assess the effectiveness of recently proposed robust training techniques and strategies, including AugMix, Revisit, and Free Normalizer, using the ResNet50 architecture on our brightness-variation dataset. Our experimental results demonstrate that these techniques can enhance the robustness of DNNs against brightness variation, leading to improved performance when dealing with images exhibiting varying brightness levels.


\end{abstract}
\section{Introduction}
Deep neural networks (DNNs) have revolutionized computer vision tasks such as image classification and object detection since the groundbreaking introduction of AlexNet in 2012 \cite{ref1}. These networks have exhibited remarkable accuracy and scalability on large-scale datasets, making them integral to a wide range of applications. However, recent investigations have unveiled the susceptibility of DNNs to intentional distortions and perturbations in input data, giving rise to concerns regarding their robustness and security. In response, researchers have embarked on a quest to develop more resilient network architectures, augmenting them with adversarial training \cite{ref2} and other robust training strategies \cite{ref38,ref39}, with the aim of fortifying the networks against potential attacks and improving their overall robustness.

The architecture of deep neural networks plays a crucial role in determining their robustness. The evolution of network architectures began with the introduction of AlexNet, one of the pioneering deep neural networks that leveraged convolutional layers to learn semantic information. AlexNet's outstanding performance, winning the top spot in the 2012 ImageNet \cite{ref3} image classification challenge and surpassing traditional machine algorithms, triggered a surge of interest in deep neural networks. Subsequent researchers focused on advancing neural network architectures. VGG \cite{ref4} introduced a modification to AlexNet by replacing the convolutional kernels with smaller ones. This not only reduced the computational cost but also improved the accuracy of the network. Szegedy et al. \cite{ref5,ref23} proposed the inception module in GoogleNet, which simulated sparse networks with dense construction. This module contributed to improved network performance. Light-weight architectures like MobileNets \cite{ref7} struck a balance between model size and accuracy, making them suitable for resource-constrained environments. ResNet \cite{ref8} introduced residual modules that addressed the challenge of learning identity maps, effectively overcoming the degradation problem in deep neural networks.

In recent years, the vulnerability of deep neural networks (DNNs) has gained significant attention \cite{ref9,ref10}. It has been observed that advanced DNNs are susceptible to adversarial examples, which are input data with subtle and often imperceptible perturbations, such as random noises and universal perturbations \cite{ref11,ref12}. However, most existing research has primarily focused on the impact of tiny, pixel-level perturbations on classification results, while paying limited attention to global, geometric, and structural transformations \cite{ref33}. In this study, we specifically investigate the impact of brightness variation in input images on the performance of deep learning models. It is widely recognized that the performance of DNNs is closely related to the data distribution of the training dataset. When there is a shift in the distribution of the testing data compared to the training data, the accuracy of DNNs tends to decline, even if the semantic information of the input image remains unchanged. Changes in the color channels of input images can alter their data distribution, leading to incorrect output predictions by the networks. Surprisingly, very few studies have explored the influence of brightness variation in images, even in common image classification tasks.

Color images contain a wealth of visual information and are commonly used in high-level computer vision tasks. The role of color in conveying information is crucial, but the underlying mechanisms of how deep neural networks perceive and process color information remain unclear. Despite the importance of color, there is a lack of research that comprehensively explains how deep networks perceive and utilize color information in their decision-making process. Recently, Kantipudi et al. \cite{ref13} conducted a notable study focusing on the impact of color channel perturbations on popular deep network architectures such as VGG, ResNet, and DenseNet. They introduced a color channel perturbation attack, where the color channels of the input images were manipulated, and assessed the resulting effect on the network's accuracy. Their findings revealed that the accuracy of these architectures decreased on average by 41\% when subjected to the color channel perturbation attack.

The current robustness benchmarking datasets like Imagenet-C, providing out-of-distribution with noise, blur, weather, cartoons, sketches distortions. Hendrycks et al. \cite{ref20} and Lau F \cite{ref21} respectively proposed a challenging dataset, Imagenet-A and NAO, which consist of real-world unmodified natural adversarial examples that most famous deep neural networks fail. As far as we know, there is few datasets designed for study the influence of brightness variation on deep networks. To help understand the impact of brightness variation, we propose an image dataset with different brightness variation on the RGB channels of images, generated from a subset of the Imagenet challenge dataset.

The main contributions of this work include the creation of a dataset related to brightness-variation images to understand their impact and then analysis the performance of most famous deep network architectures on image classification task on the proposed dataset. Then we use ResNet50 as an example to study the influence of robust learning techniques like Augmix \cite{ref38}, Revisiting \cite{ref39} and Normalizer Free \cite{ref40} on networks’ capability to resist the impact of brightness variation. Finally, we examine the relation between the robustness of DNNs on brightness variation and their depth. The rest of the paper is organized as follows: Section \ref{sec2} presents background information related to the existing literature, and Section \ref{sec3} presents how we constructed the dataset and provides details of the experiments, followed by the results and findings and finally the conclusion and some discussions are given in Section \ref{sec4}.

\section{Background}
\label{sec2}

In this section we present a review of relevant literature as well as some background knowledge. 


The quality of training data and the effect of perturbations on deep neural networks (DNNs) are crucial aspects that have been extensively studied in recent years. Researchers have discovered that DNNs learn both feature representations and semantic information from the data distribution in their training data. However, when the images are perturbed by geometric transformations, deletions, blurring, or other distortions, the data distribution changes, which can lead to misclassification by the neural network. The concept of adversarial attacks was introduced by Szegedy et al. \cite{ref6}, highlighting the phenomenon of error amplification rather than nonlinearity or overfitting as the reason behind the success of attacks. Dodge and Karam \cite{ref14} investigated the impact of adversarial samples on DNN performance and explored how different types of distortions and perturbations affect the classification paradigm of DNNs. Their experiments involved the use of the Imagenet dataset and models such as Caffe Reference \cite{ref22}, VGG16 \cite{ref4}, and GoogleNet \cite{ref5}. To improve the robustness of DNNs against various types of distortions, Dodge and Karam \cite{ref24,ref25} proposed ensemble methods based on a mixture of experts. This approach combines multiple expert models in a weighted manner to enhance the network's resilience. Additionally, they compared the classification performance of humans and DNNs on distorted images. The findings showed that humans generally outperform DNNs in classifying distorted images, even when the DNNs are retrained with distorted data.

Zhou et al. \cite{ref28} demonstrated in their work that by fine-tuning and re-training DNNs, their performance in classifying distorted images can be enhanced. This highlights the importance of adapting the network to the specific characteristics of the perturbations.
Borkar and Karam \cite{ref26} proposed a criterion to evaluate the impact of perturbations, such as Gaussian blur and additive noise, on the activations of pre-trained convolutional filters. By ranking the most noise vulnerable convolutional filters in commonly used convolutional neural networks (CNNs), they aimed to identify the filters that could benefit the most from correction to achieve the highest improvement in classification accuracy.
Hossain et al. \cite{ref27} conducted an analysis of the performance of VGG16 when influenced by various types of perturbations, including Gaussian white noise, scaling Gaussian noise, salt $\&$ pepper noise, speckle, motion blur, and Gaussian blur. To improve the network's robustness against these distortions, they employed discrete cosine transform during the training process.

\subsection{Impact of Colour}

Dosovitskiy and Brox \cite{ref15} were the first to show that manipulating the color of an object in a way that deviates from the training data has a negative impact on classification performance. Engilberge et al. \cite{ref16} managed to identify the colour-sensitive units that processed hue characteristics in the VGG-19 and AlexNet.

Kantipudi et al. \cite{ref13} proposed a colour channel perturbation attack to fool deep networks and defense it by data augmentation. Le and Kayal \cite{ref17} compared various models to show that the robustness of edge detection is an important factor contributing to the robustness of models against color noise. Kanjar et al. \cite{ref18} analyzed the impact of colour on robustness of widely used DNNs. They performed experiments on their proposed dataset with hue space based distortion. Hendrycks et al. \cite{ref22} have also used the validation set of the Imagenet database as the base database and augmented different colour distorted images from these images. Their works inspire us to further study the impact of brightness variation on DNNs.

\subsection{Architectures} 
Alexnet \cite{ref1} is one of the origins of most common neural network architectures. It pioneered to take the use of GPUs to accelerate the training of neural networks, reducing the training time of the neural network to an acceptable range. The success of AlexNet on Imagenet has motivated more work on the development of DNNs’ architecture. VGG \cite{ref4} took advantage of Alexnet and used several consecutive $3 \times 3$ convolution kernels instead of the larger ones in AlexNet ($11 \times 11$, $7 \times 7$, $5 \times 5$) to improve the performance. They pointed out that for a given receptive field, using consecutive smaller convolution kernels is better than a larger convolution kernel, as it makes the network deeper and more efficient. Another key innovation in deep network architectures is the inception module, which consists of $1 \times 1$, $3 \times 3$ and $5 \times 5$ convolutions. Inception approximate and covers the best local sparse structure of the convolutional network by easily accessible dense components. With the rapid development of the architecture of deep networks, the models have become much deeper, which caused vanishing gradients and degradation problems. The ResNet \cite{ref8} architecture was proposed to tackle these issues, and is still one of the widely used backbones in computer vision tasks nowadays. ResNet proposed a structure called residual block, which skips connections between adjacent layers, enabling the network to learn identity mapping easier. It ensures that the deeper network at least perform as good as smaller ones.  Based on ResNet architecture, DenseNet \cite{ref41} proposes a more radical intensive connectivity architecture, which connects all the layers to each other and each layer takes all the layers before it as its input. DenseNet needs fewer parameters compared to the other traditional convolutional DNNs by reducing the need to learn redundant features. As the networks become deeper, their drawbacks of being computationally intensive and require a lot of GPU memory become more significant, which makes them unsuitable for mobile devices. To deploy models in such portable devices a group of lightweight networks were proposed, and Mobilenet \cite{ref7} is one of the most famous architectures among them. It proposed the concepts of depth wise separable convolutions and inverted residuals, which achieve similar performance to traditional networks with less computational cost. 

\subsection{Robustness} 
Evaluating robustness of deep neural networks is still a challenging and ongoing area of research \cite{ref29}. Some works use data augmentation to improve the robustness of networks \cite{ref34}. Papernot \cite{ref35} et al. first pointed out some limitations of deep learning in adversarial settings. They proposed forward derivative attack to fool DNNs by only alter a minority of input. Hendrycks et al. \cite{ref36} defined some benchmark metrics of robustness of DNNs to some common perturbations like additive noise, blur, compression artifacts, etc. They proposed a variant of Imagenet referred to as Imagenet Challenge (Imagenet-C). Imagenet-C contains 15 types of automated generated perturbations, on which many well-known DNNs perform poorly. Kanjar et al. \cite{ref18} analyzed the impact of colour on robustness of widely used DNNs. Recent studies have indicated that convolutional DNNs pre-trained on Imagenet dataset are vulnerable to texture bias \cite{ref38}, while the impact of scaling in images is not deeply studied. Xiao et al. \cite{ref31} have formalized the scaling attack, illustrating its goal, generation algorithms, and optimization solution. Zheng et al. \cite{ref32} provide a basis for robustness evaluation and conduct experiments in different situations to explore the relationship between image scaling and the robustness of adversarial examples. With the development of  adversarial attack techniques, many studies focus on defensing against these attacks and try to find feasible training strategies to improve the robustness of models \cite{ref19}. Augmix \cite{ref38} is a simple training strategy, which uses several augmentation techniques together with Jenson-Shannon divergence loss to enforce a common embedding for the classifier. Brock et al. \cite{ref40} proposed a normalized family of free networks called NF-Nets to prevent the gradient explosion by not using batch normalization. Tan et al. \cite{ref39} recently showed that network models can effectively improve the classification performance of ResNet models by using some scaling strategies and developed a set of models called ResNet-RS.

\begin{figure}[t]
\centering
\includegraphics[width=0.9\columnwidth]{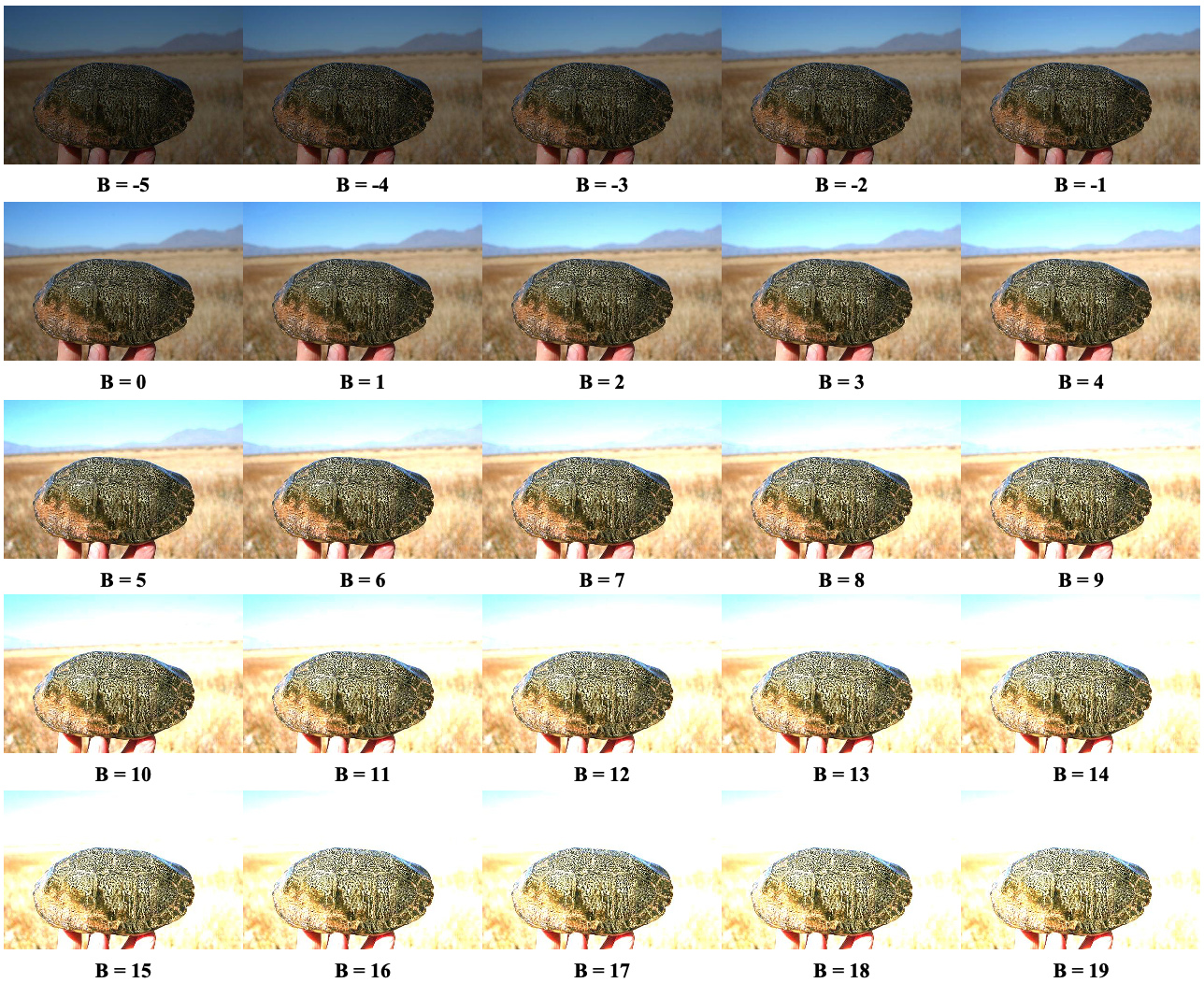} 
\caption{Examples of generating brightness-variation images.}.
\label{figure1}
\vspace{-5pt}
\end{figure}

\section{Methodology and experiments}
\label{sec3}
\subsection{Dataset generation}
Typically, DNNs trained on Imagenet dataset have 1000 different labels. Our proposed dataset is derived from a subset of Imagenet Challenge dataset, and we call it ImageNet-BrightnessVariation (ImageNet-BV). Firstly, 50 images were randomly selected from each category of Imagenet Challenge to generate a clean sample dataset, which totally contains 50,000 raw images. Secondly, the whole dataset of 1,200,000 images is generated by adjusting 24 different brightness levels for each image. As shown in Figure \ref{figure1}, one original image ($B = 0$) is augmented into 24 images with different brightness levels.  Here, our method of generating noise samples is expressed as follows:

\begin{equation}
    \label{Formula 1}
    {X}_{B}=Clip(X \otimes (1+B/10))
\end{equation}
where $X$ represents the clean sample, ${X}_{B}$ represents the noise sample, $X \otimes n$ represents the value of R, G and B channels of $X$ is multiplied with $n$, $Clip$ represents the value greater than 255 or less than 0 is set to 255 and 0, respectively.

\subsection{Impact on widely used network architectures}

In this section, we conduct experiments to evaluate the accuracy of six widely utilized deep network architectures \cite{ref41,ref8,ref4,ref5,ref7,ref1} using the generated ImageNet-BV dataset. The classification accuracy of the images without any brightness perturbation is represented by the sixth row ($B=0$), serving as a reference for comparison.

\begin{table}[t]
	\centering
    \setlength{\belowcaptionskip}{10pt}
    \caption{Classification Top-1 accuracy (\%) of well-known DNNs on ImageNet-BV.}.
    \label{Table1}
	\begin{tabular}{ccccccc}

    \hline
    $B$ & DenseNet & ResNet50 & VGG19 & GoogleNet & MobileNet & AlexNet\\
    \hline
    -5&78.99&79.99&76.14&72.24&73.85&58.34\\
	\hline
    -4&80.99&82.79&79.06&74.21&76.71&63.67\\
	\hline
    -3&81.81&84.17&80.51&75.11&78.58&67.32\\
	\hline
    -2&82.08&84.98&81.39&75.71&79.60&69.56\\
	\hline
    -1&82.06&85.49&82.00&\textbf{76.01}&80.35&70.86\\
	\hline
 0&\textbf{82.15}&\textbf{85.75}&\textbf{82.52}&75.85&\textbf{80.70}&\textbf{71.61}\\
	\hline
 1&81.56&85.28&82.03&75.60&80.13&70.97\\
	\hline
 2&80.89&84.39&81.01&75.02&79.13&69.97\\
	\hline
 3&79.90&83.28&79.60&74.34&77.83&68.27\\
	\hline
 4&78.61&82.16&78.09&73.53&76.40&65.99\\
	\hline
 5&77.32&80.57&76.19&72.78&74.53&63.49\\
	\hline
 6&76.03&78.95&74.25&71.64&72.77&60.56\\
	\hline
 7&74.38&77.24&72.23&70.35&70.79&57.69\\
	\hline
 8&72.81&75.50&70.03&69.02&68.76&54.64\\
	\hline
 9&71.10&73.51&67.76&67.69&66.61&51.51\\
	\hline
 10&69.31&71.64&65.62&66.45&64.44&48.60\\
	\hline
 11&67.51&69.74&63.30&64.84&62.38&45.88\\
	\hline
 12&65.55&67.65&60.99&63.33&60.27&43.21\\
	\hline
 13&63.64&65.68&58.57&61.94&58.07&40.79\\
	\hline
 14&62.02&63.46&56.29&60.41&56.00&38.33\\
	\hline
 15&60.11&61.59&53.90&58.85&53.95&35.84\\
	\hline
 16&58.41&59.60&51.49&57.26&51.94&33.77\\
	\hline
 17&56.63&57.64&49.33&55.83&49.99&31.95\\
	\hline
 18&54.89&55.83&47.34&54.44&48.06&30.17\\
	\hline
 19&53.21&53.92&45.39&53.00&46.31&28.52\\
	\hline

\end{tabular}
\end{table}

The experimental results presented in Table \ref{Table1} provide several key findings:

\textbf{1)} Various levels of brightness variation consistently lead to a decrease in the top-1 classification accuracy of advanced deep neural networks (DNNs). This observation holds true across all evaluated network architectures; \textbf{2)} Upon closer examination of the influence of different architectures, ResNet50 demonstrates the highest classification accuracy among the models. DenseNet and VGG19 exhibit similar performance in terms of classification accuracy, surpassing MobileNet and GoogleNet. Notably, AlexNet consistently performs the worst in terms of classification accuracy across all brightness variations; \textbf{3)} With the exception of GoogleNet, a clear pattern emerges in the classification accuracy of DNNs with respect to brightness variation: When $B$ is negative ($B<0$), the classification accuracy increases as $B$ increases; when $B$ is zero ($B=0$), the accuracy reaches its maximum; when $B$ is positive ($B>0$), the accuracy decreases as $B$ increases.

These observations provide valuable insights into the impact of brightness variation on DNNs, highlighting the varying performance of different architectures and the influence of brightness levels on classification accuracy.

It indicates that although the visual information of the images is barely changed, the brightness variation on images still has significant impact on the performance deep neural networks. Despise the tested DNNs all contains convolutional layers, which gives them the property of geometric invariance, they are still vulnerable to brightness variation.

\begin{figure}[t]
\centering
\includegraphics[width=1\columnwidth]{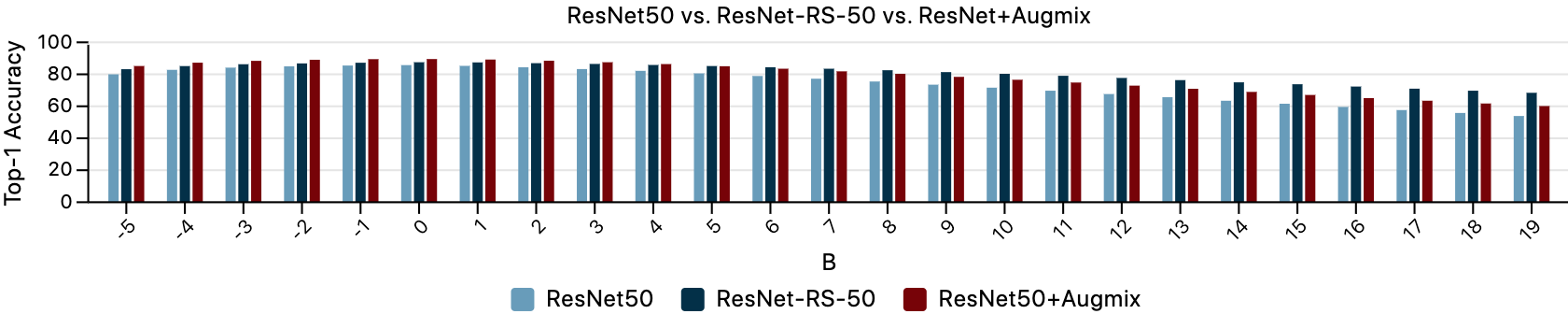} 
\caption{Performance of ResNet50 vs. ResNet-RS-50 vs. ResNet50+Augmix.}.
\label{figure2}
\vspace{-10pt}
\end{figure}

\subsection{Recent advances in efficient and robust models}

\textbf{Augmix and ResNet-RS-50.} Hendrycks et al. \cite{ref38} introduced AugMix as a data processing technique aimed at enhancing the robust performance of deep neural networks. AugMix employs various augmentation methods, such as rotation, translation, and channel mixing, to augment the input data. In our experiments, we evaluate the classification performance of ResNet50 with AugMix data processing technology using our dataset. We compare the classification performance of a pretrained ResNet50 model with that of a pretrained ResNet50 model with AugMix technology, and the results are illustrated in Figure \ref{figure2}. The findings reveal that when input images with brightness variation are utilized, the ResNet50 model with AugMix technology achieves better classification performance compared to the ordinary pretrained ResNet50 model. Additionally, Bello et al. \cite{ref39} recently demonstrated the effectiveness of scaling network models in improving classification performance. They developed a set of models known as ResNet-RS. Here, we present the classification performance of ResNet-RS-50 on Imagenet-CV, with its Top-1 accuracy depicted in Figure \ref{figure2}. The results clearly indicate that ResNet-RS-50 exhibits superior robustness compared to the pretrained ResNet50 model.

The findings demonstrate the effectiveness of AugMix in enhancing the classification performance of ResNet50 and the improved robustness of ResNet-RS-50, particularly when faced with brightness variation in input images. In general, both ResNet50 with AugMix and ResNet-RS-50 exhibit greater robustness compared to the original pretrained ResNet50 model. Furthermore, Figure \ref{figure2} provides additional insights: \textbf{1)} Similar to ResNet50, the Top-1 classification accuracy of ResNet-RS-50 and ResNet+AugMix reaches its maximum when $B=0$; \textbf{2)} As the level of distortion added to the color channels of images increases, the classification accuracy of ResNet50 with AugMix and ResNet-RS-50 also decreases accordingly. This observation indicates that brightness still possesses a certain level of antagonism towards the improved ResNet50 models; \textbf{3)} When $B<=4$, ResNet-RS-50 exhibits less robustness compared to ResNet+AugMix. However, when $B>4$, ResNet-RS-50 showcases greater robustness compared to ResNet+AugMix.

\begin{figure}[t]
\centering
\includegraphics[width=1\columnwidth]{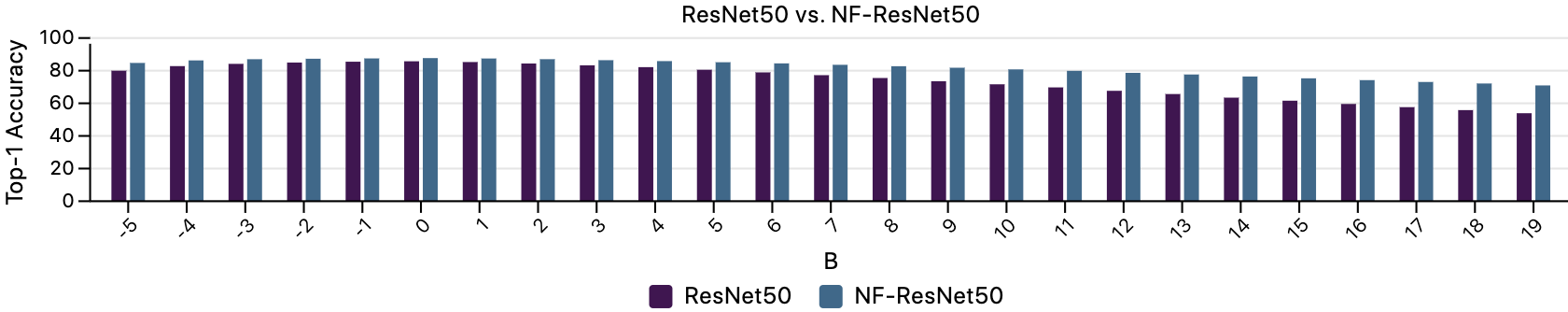} 
\caption{Performance of ResNet50 vs. NF-ResNet50.}.
\label{figure3}
\vspace{-10pt}
\end{figure}

\textbf{Normalizer Free ResNet50(NF-ResNet50).} Brock et al. \cite{ref40} introduced a family of normalized free networks called NF-Nets, which differ from traditional networks by not employing batch normalization. During the training of NF-Nets, measures are taken to limit the size of gradients, effectively preventing gradient explosion and promoting training stability. Figure \ref{figure3} illustrates the Top-1 accuracy of NF-ResNet50 on our dataset, revealing the following observations: \textbf{1)} NF-ResNet50 exhibits improved classification performance compared to the pretrained ResNet50 model. This enhancement highlights the effectiveness of the NF-Net architecture; \textbf{2)} Similar to ResNet50, NF-ResNet50 achieves its highest Top-1 classification accuracy when $B=0$. As additional distortions are introduced to the color channels of images, the classification accuracy of NF-ResNet50 decreases correspondingly. This finding suggests that brightness still exerts a certain antagonistic effect on the performance of NF-ResNet50. These results provide insights into the performance characteristics of NF-ResNet50, showcasing its improved classification performance compared to the pretrained ResNet50 model, as well as its sensitivity to brightness variations in input images.

\section{Conclusion and future work}
\label{sec4}

The widespread usage of deep neural networks in various applications necessitates a thorough examination of their robustness and the development of techniques to enhance their resistance to perturbations. This work presents experimental studies that shed light on the impact of brightness variation on the performance of deep neural network architectures, particularly in relation to shifts in data distribution. The findings reveal a significant reduction in the performance of these networks as distortions are introduced to the color channels of images.

The study also investigates the effectiveness of data processing and augmentation techniques, such as AugMix and Revisit, in improving the robustness and optimization of deep network training. Notably, when considering brightness variation, ResNet50+AugMix and ResNet-RS-50 demonstrate greater robustness compared to ResNet-50. Furthermore, the study explores the performance of Normalizer free models, which have been found to exhibit enhanced robustness to brightness variation compared to pretrained ResNet50 models.

These observations underscore the substantial impact of brightness variation on the inference of advanced deep neural networks. The positive influence of data processing and augmentation techniques highlights the potential for further improving the robustness and accuracy of deep neural network models. These findings will serve as a driving force for future research on the brightness perception mechanisms of other architectural studies. The insights gained from this analysis will motivate researchers to consider the influence of brightness and other perturbations in the development of more accurate and robust deep neural network models.


\bibliographystyle{splncs04}
\bibliography{prcv}

\end{sloppypar}
\end{document}